\documentclass[11pt,a4paper]{article}
\usepackage[hyperref]{acl2021}
\usepackage{times}
\usepackage{latexsym}

\usepackage{microtype}

\usepackage{graphicx}
\usepackage{caption}
\usepackage{booktabs}
\usepackage{multirow}
\usepackage{amsmath}
\usepackage{amssymb}
\usepackage{enumitem}
\usepackage{soul}
\usepackage[switch]{lineno}
\definecolor{lightblue}{rgb}{.90,.95,1} 
\newcommand{\furl}[1]{\footnote{\url{http://#1}}}
\definecolor{darkblue}{rgb}{0, 0, 0.5}
\definecolor{pastelred}{rgb}{1.0, 0.41, 0.38}
\definecolor{lightgreen}{rgb}{0.56, 0.93, 0.56}
\newcommand{\hlred}[1]{{\sethlcolor{pastelred}\hl{#1}}}
\newcommand{\hlgreen}[1]{{\sethlcolor{lightgreen}\hl{#1}}}

\aclfinalcopy %

\title{Medical Code Assignment with Gated Convolution and Note-Code Interaction}

\author{
        Shaoxiong Ji~\textsuperscript{\dag},~
        Shirui Pan~\textsuperscript{\ddag},~
        Pekka Marttinen~\textsuperscript{\dag} \\
    \textsuperscript{\dag} Department of Computer Science, Aalto University \\
    \textsuperscript{\ddag} Faculty of Information Technology, Monash University \\
    \texttt{\{shaoxiong.ji;~pekka.marttinen\}@aalto.fi} \\
    \texttt{shirui.pan@monash.edu}
}

\date{}

\begin{document}
\maketitle
\begin{abstract}
Medical code assignment from clinical text is a fundamental task in clinical information system management. As medical notes are typically lengthy and the medical coding system's code space is large, this task is a long-standing challenge. 
Recent work applies deep neural network models to encode the medical notes and assign medical codes to clinical documents. However, these methods are still ineffective as they do not fully encode and capture the lengthy and rich semantic information of medical notes nor explicitly exploit the interactions between the notes and codes. We propose a novel method, gated convolutional neural networks, and a note-code interaction (GatedCNN-NCI), for automatic medical code assignment to overcome these challenges. Our methods capture the rich semantic information of the lengthy clinical text for better representation by utilizing embedding injection and gated information propagation in the medical note encoding module. With a novel note-code interaction design and a graph message passing mechanism, we explicitly capture the underlying dependency between notes and codes, enabling effective code prediction. A weight sharing scheme is further designed to decrease the number of trainable parameters.
Empirical experiments on real-world clinical datasets show that our proposed model outperforms state-of-the-art models in most cases, and our model size is on par with light-weighted baselines. 
\end{abstract}

\section{Introduction}
\label{sec:introduction}

Automatic medical code assignment is a routine healthcare task for medical information management and clinical decision support.
The International Classification of Diseases (ICD) coding system, maintained by the World Health Organization (WHO), is widely used among various coding systems. 
Thus, the medical code assignment task is also called ICD coding.
It uses all types of clinical notes to predict medical codes in a supervised manner with human-annotated codes~\cite{perotte2014diagnosis}, which is formulated as a multi-class multi-label text classification problem in the medical domain.

While there are increasing works in the community in automatic medical code assignment~\citep{prakash2017condensed, shi2017towards, mullenbach2018explainable, ji2020dilated}, this task remains challenging from the perspectives of note representation and code prediction.
First, medical note representation, a critical step in understanding medical notes, is formidably challenging due to the lengthy and complex semantic information in the discharge documents. There are typically thousands of tokens in a medical note due to the various diagnoses and procedures experienced by a patient. Furthermore, clinical notes also contain a vocabulary with many professional words and phrases, making it hard for a neural network model to encode and understand critical information.
Second, the medical coding system has a very high and sparse dimensional label space, which renders the code prediction task incredibly difficult.
For example, ICD9 and ICD10 coding systems have many labels, i.e., more than 14,000 and 68,000 codes. However, a patient typically is diagnosed with only a couple of codes over the whole coding space.

\begin{figure*}[htbp!]
\centering
	\includegraphics[width=0.9\textwidth]{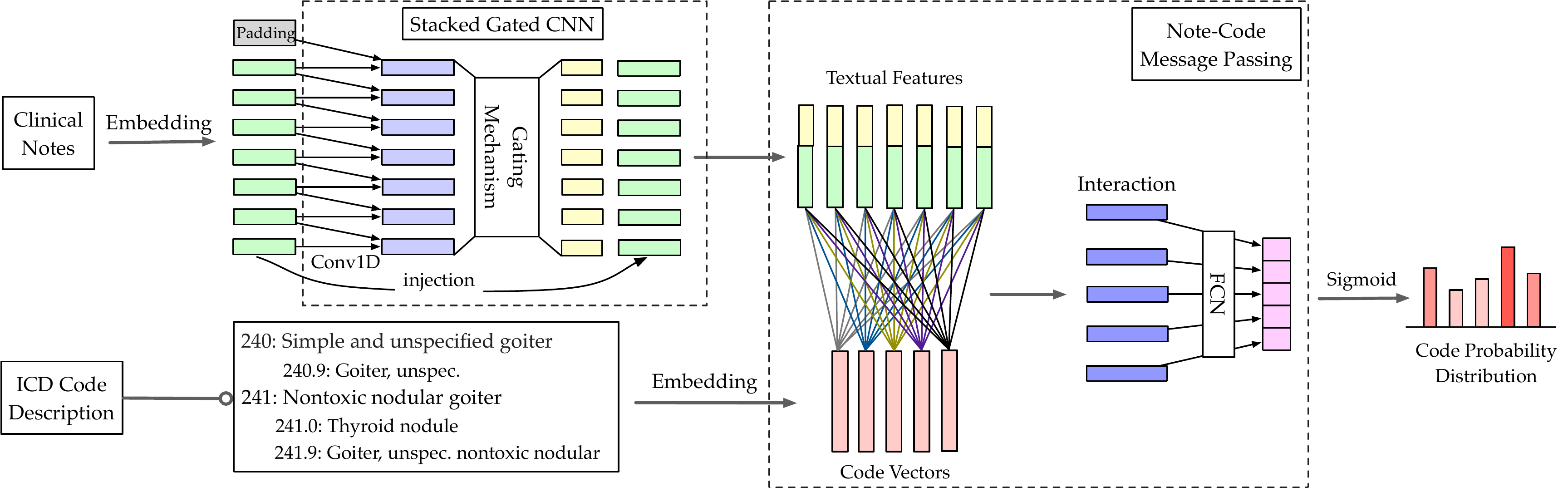}
	\caption{Illustration of the GatedCNN-NCI model architecture. The gating mechanism controls the information propagation. Textual features interact with each code vector in the note-code interaction module. FCN is a fully connected layer.}
	\label{fig:model}
\end{figure*}

Early works for medical code assignment typically follow statistical approaches. They either employ rule-based methods \cite{farkas2008automatic} or apply classification methods such as SVM and Bayesian ridge regression \cite{lita2008large} to assign the codes. These methods are shallow and do not exploit the complex semantic information in medical notes, leading to unsatisfactory performance. Recently, Natural language processing (NLP) techniques based on deep learning have been developed \cite{mullenbach2018explainable, li2018automated, cao2020hypercore, ji2020dilated}, which learn the note representation via convolutional neural networks. Specifically, CAML~\cite{mullenbach2018explainable}, MultiResCNN~\cite{li2018automated} and DCAN~\cite{ji2020dilated} treat ICD coding as a general text classification problem and develop complex neural encoders to learn the note representation. HyperCore~\cite{cao2020hypercore} proposes the hyperbolic embedding to capture code hierarchy and co-occurrence.
However, these approaches are still ineffective, as they do not explicitly capture the fine-grained interactions between textual elements and medical codes. These interactions naturally represent the interdependencies between the complex medical words and associated codes, and thus should be well exploited.

This paper puts forward a novel neural architecture,  Gated Convolutional Neural Network with Note-Code Interaction (GatedCNN-NCI), for effective medical code assignment. Our goal is to learn rich representation from clinical notes and exploit the interactions between medical texts and clinical codes. To capture the long sequential history of clinical documents, we design a novel dilation information propagation component with a forgetting mechanism to selectively utilize the useful information for note representation learning. To tackle the large labeling space, we formulate textual notes and medical codes as a complete bipartite graph and develop a graph message passing approach to capture the explicit interaction between notes and codes. The ICD code descriptions are used as an external medical knowledge source to learn more accurate code representations that preserve the semantic relations of the codes.
Considering the practical application in real-world medical institutes, especially those with limited computing resources, our architecture also prioritizes computational efficiency when designing the sub-modules. 
Our contributions are itemized as follows.

\begin{itemize}[noitemsep]
\item We propose a CNN-based neural architecture with dilation and gating mechanism for clinical text encoding. We enhance the feature representation learning with 1) embedding injection, enhancing the deeper features of lengthy clinical notes; 2) and the gating mechanism to control the information propagation. 
\item We view the note-code interaction as a complete bipartite graph and propose a graph message passing mechanism to capture the interactions between textual features and ICD codes explicitly. 
\item To reduce the trainable parameters and make our model computationally efficient, we develop a weight-sharing mechanism across the length of the sequence and the depth of the network.  
\item Experiments in real-world clinical datasets empirically validate our model's effectiveness by comparison with the state of the art.  
\end{itemize}

\section{Related~Work}
\label{sec:related}

Classical medical coding systems used rule-based methods~\citep{farkas2008automatic}, studied feature selection~\citep{medori2010machine}, and applied classification models such as SVM and Bayesian ridge regression~\citep{lita2008large}.
~\citet{perotte2014diagnosis} utilized the hierarchical structure of the ICD code systems and provided a flat and hierarchical SVM for diagnosis code classification,
while
~\citet{kavuluru2015empirical} studied explicit co-occurrence relations between codes. 
\citet{scheurwegs2016data} investigated heterogeneous data of both structured records and textual data. 
Recent deep learning-based models use word embedding techniques and develop complex neural network architectures to learn rich text features for automatic medical code assignment.
Popular models use recurrent architectures such as the LSTM network with an attention mechanism~\citep{shi2017towards} and GRU network with hierarchical attention~\citep{baumel2018multi}.
\citet{prakash2017condensed} used Wikipedia as a knowledge source and proposed condensed memory networks (C-MemNNs) with iterative condensation of memory representation.
Although CNNs are traditionally applied in computer vision, many ICD coding methods utilize convolutional architectures. 
CAML~\citep{mullenbach2018explainable} used CNN with multiple filters and label attention. 
\citet{li2018automated} adopted the doc2vec embedding and CNN architecture, and \citet{bai2019improving} incorporated online knowledge sources.
The recent MultiResCNN model~\citep{li2020multirescnn} extensively concatenated and stacked CNNs with multi-filter convolution and residual learning. 
HyperCore~\cite{cao2020hypercore} utilized hyperbolic embedding and co-graph representation to capture the code hierarchy.

\section{Method}
\label{sec:method}

\subsection{Problem Definition}
The input clinical note with $n$ words is denoted as $\mathbf{x}_{1: n}=x_{1}, \ldots, x_{n}$, where each $x_i$ is a word (or token).
The medical coding system is the set of all possible diagnosis and procedure codes denoted as $\mathcal{C}$. 
The medical code assignment learns a function $\mathcal{F}: \mathcal{X}^{n} \rightarrow \mathcal{Y}^m$ such that 
\begin{equation}
y=\mathcal{F}\left(x_{1}, \ldots, x_{n}; \mathcal{D}\right),
\end{equation}
where $y\in \mathbb{R}^m$ is the medical code at discharge, $m$ is the number of medical codes, and $\mathcal{D}$ is an optional external knowledge source. 
This paper uses the ICD coding system and naturally utilizes the official textual ICD code description as an external knowledge source. 

\subsection{High-level Model Architecture}
The high-level model architecture of GatedCNN-NCI is illustrated in Fig.~\ref{fig:model}.
Our model consists of two main components, i.e., stacked gated CNN layers for clinical note encoding and note-code interaction to fuse the external ICD code description.  
The stacked gated CNNs include three sub-modules, i.e., dilated convolution, embedding injection, and gating mechanism. 

We use word2vec~\citep{mikolov2013distributed} to train word embeddings from raw tokens. 
Word embedding matrix of a clinical note is denoted as $[ \mathbf{w}_1, \dots, \mathbf{w}_n]^{\operatorname{T}}\in \mathbb{R}^{n\times d_e}$, where $d_e$ is the dimension of word vectors. 
Then we input the word embeddings into stacked gated CNN layers for long-range information propagation. 
The stacked module uses dilated convolution as its backbone~\cite{oord2016wavenet}. 
To further enhance the feature learning, we inject the original embedding into each stacked layer. 
The gating mechanism is originated from the long short-term memory network (LSTM)~\cite{hochreiter1997long}.  
We adopt the LSTM-like gate~\cite{dauphin2017language} to control the information flow. 

To avoid blurry memory in higher layers, we inject the original word embeddings~\cite{bai2019trellis}.    
Label interaction has been studied by~\citet{wang2016learning} and \citet{du2019explicit}.  
We utilize descriptive knowledge from the ICD code descriptions and develop the note-code interaction to capture the relational match between clinical note features and ICD codes. 
To reduce the training cost and stabilize the training process, we also introduce a weight sharing mechanism across the stacked CNNs~\cite{bai2019trellis}.

\subsection{Dilated Convolutional Layers}
We use the one-dimensional convolution with dilation as the backbone of our encoder, which takes the word embedding $\mathbf{X}\in \mathbb{R}^{n\times d_e}$ as input. 
Dilated CNN has exhibited a significant capacity for long sequence modeling and computationally efficient for parallelism~\cite{bai2018empirical}.
Specifically, we use a 1D convolution operator $\operatorname{Conv1D}(x ; f)$, with a filter $f:\{0, \ldots, k-1\} \rightarrow \mathbb{R}$, to each dimension of the word vectors. 
Given a sequence of one-dimensional elements $\mathbf{x}\in \mathbb{R}^n$, the one-dimensional dilated convolution $\mathcal{F}_d$ is denoted as 
\begin{equation}
\mathcal{F}_d(s)=\left(\mathbf{x} *_{d} f\right)(s)=\sum_{i=0}^{k-1} f(i) \cdot \mathbf{x}_{s-d \cdot i},
\end{equation}
where $d$ is the dilation size (i.e., the space between kernel elements), $s$ is the index of the element of the input sequence, $k$ is the convolving kernel (aka, the filter) size, and $s-d\cdot i$ refers to past time steps. The dilation size of $d$ and kernel size $k$ control the receptive field. 
The 1D dilated convolution has $d_h$ output channels, i.e., for each of the $d_e$ input channels $d_h$ convolutional features are learned through the dilated $\operatorname{Conv1D}$. 
Stacking CNN layers can be adopted to learn in-depth features.

\subsection{Embedding Injection}
Our hypothesis for encoding a very long clinical sequence is that the deep neural encoding architecture tends to forget important information, mainly because the clinical note contains fruitful professional expression about the patient's diagnosis. 
Thus, in-depth features become blurry with the increase of neural layers. 
We propose to inject original word embedding into each intermediate layer of the proposed architecture, attempting to remind the network to reactivate the original diagnostic notes and mitigate the failure of extracting meaningful, in-depth features. 
We denote the hidden representation at the $l$-th layer as $\mathbf{H}^l\in \mathbb{R}^{n\times d_h}$, where the dimension $d_h$ is the hidden dimension.
Word embedding is concatenated into $l$th-layer hidden representation as
\begin{equation}
	\mathbf{J}^l = \operatorname{concat}\lbrack \mathbf{X}, \mathbf{H}^l \rbrack,
\end{equation}
where $\mathbf{J}^l\in \mathbb{R}^{n\times (d_e+d_h)}$ are the deep features enhanced with the original clues, used as the new input of the next convolutional encoding layer. 
We randomly initialize the $\mathbf{H}^0$ matrix for the first convolutional layer.    

\subsection{Gating Mechanism}
\begin{figure}
\centering
	\includegraphics[width=0.35\textwidth]{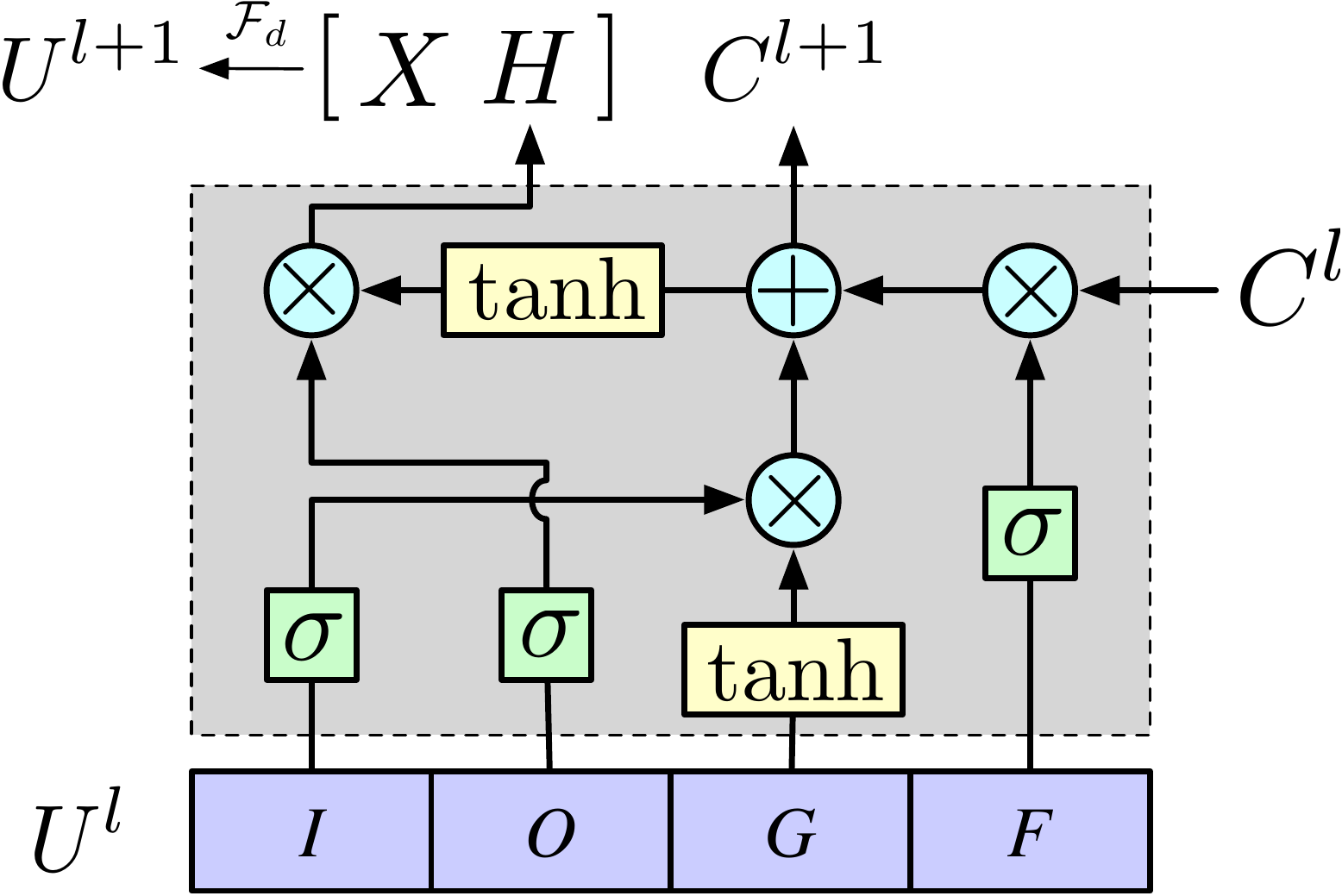}
	\caption{Gating mechanism that controls the flow's convolutional features through stacked layers' depth.}
	\label{fig:gate}
\end{figure}

Embedding injection of original word vectors brings low-level features to higher-level, which may lead to difficulty in feature learning in higher layers.  
Thus, we develop an LSTM-style gating mechanism to control the information flow and capture a long history in the sequence. 
Unlike the recurrent gate such as the LSTM that controls the information flow along the time coordinate, this gating mechanism controls the flow through stacked layers' depth. 
The gating mechanism is depicted in Fig.~\ref{fig:gate}, where $\sigma$ and $\operatorname{tanh}$ are sigmoid and hyperbolic tangent activation functions respectively.
After the embedding injection, the dilated CNN upsamples the injected signal $\mathbf{J}^l$ into $\mathbf{U}^l\in \mathbb{R}^{n\times d_u}$ at the $l$-th layer.
We divide $\mathbf{U}^l$ into four matrices with the same dimension, i.e., $\mathbf{I}$, $\mathbf{O}$, $\mathbf{G}$ and $\mathbf{F} \in \mathbb{R}^{n\times d_g}$, such that: 
\begin{equation}
	\mathbf{U}^l = \operatorname{concat}\lbrack \mathbf{I}, \mathbf{O}, \mathbf{G}, \mathbf{F} \rbrack.
\end{equation}
Here, we have $d_u=4\times d_g$.
Then, these four matrices are fed into the LSTM-like gating module that controls what information should be propagated to deeper layers. 
The input gate $\sigma(\mathbf{I})$ decides the information to be infused and stored into the cell state $\mathbf{C}$. 
The forget gate $\sigma(\mathbf{F})$ chooses the information to be remembered. 
The output gate $\sigma(\mathbf{O})$, working with the cell state, focuses on what signals propagate into the next layer.
This process is formalized as 
\begin{gather*}
\mathbf{C}^{l+1} = \sigma(\mathbf{F}) * \sigma(\mathbf{C}^l) + \sigma(\mathbf{I})* \mathop{tanh}(\mathbf{G}) \\
\mathbf{H} = \sigma(\mathbf{O}) * \mathop{tanh}(\mathbf{F}),
\end{gather*}
where $\mathbf{C}^l$ is the cell state at the $l$-th layer and $\mathbf{H}$ is the hidden state produced by the gated unit. 
The embedding injection trick concatenates the original word embedding $\mathbf{X}$ and the hidden representation $\mathbf{H}$, and the dilated convolutional layer upsamples the concatenation to get the new feature $\mathbf{U}^{l+1}$ at the $(l+1)$-th layer, denoted as: 
\begin{equation}
	\mathbf{U}^{l+1} \stackrel{\mathcal{F}_{d}}{\longleftarrow}\lbrack \mathbf{X}, \mathbf{H}\rbrack.
\end{equation}

Gated CNNs can be stacked into a deep architecture, as shown in the general framework of Fig.~\ref{fig:model}. 
As a result, our model can represent a large-sized context and extract hierarchical features at each layer. 
Moreover, the gating mechanism can also extract important features to remember and focus, while less critical features are forgotten and ignored at each layer.

\subsection{Note-Code Interaction as Message Passing}
To capture the explicit note-code interaction (NCI) between the medical codes and textual mentions, we build a complete bipartite graph $G=\lbrace U, V, E\rbrace$, where $U=\{w_i\}^n$ and $V=\{c_j\}^m$ represent the words and ICD codes respectively, and $E$ is the fully connected edge set.
For simplicity, we omit the superscript of the last convolutional features $\mathbf{U}^{l+1}$ extracted by the stacked gated CNNs and denote the textual node features $\mathbf{U}$ as the vertex set $U$ in the note-code bipartite graph.
We incorporate the ICD code descriptions of WHO to represent the medical knowledge about ICD codes.
For example, the ICD code 240 in Fig.~\ref{fig:model} is about simple and unspecified goiter.
Instead of merely using the ICD code index to represent the prediction target, we include the code description, which contains rich domain knowledge. 
Word embeddings of description are averaged to obtain code vectors $\mathbf{V}\in \mathbb{R}^{m\times d_v}$, where $m$ is the number of codes, and $d_v$ is the embedding dimension. 
We take the code vectors as the node features of the vertex set $V$.

Our novel formulation of the bipartite graph preserves the source-target matching between textual features and ICD code vectors.
We utilize the graph message passing mechanism~\cite{gilmer2017neural, wu2020comprehensive} to infer fine-grained clues about dependencies between textual features and code semantics. 
The composition function $\operatorname{NCI}: \mathbb{R}^{n\times d_u} \times \mathbb{R}^{m\times d_v}\rightarrow \mathbb{R}^{m}$ is denoted as:
\begin{equation}
	\operatorname{NCI}(U, V) = f_\theta \bigg ( \sum_{i,j}g_{\xi}(w_i, c_j) \bigg ),
\end{equation} 
where $g_\xi$ with parameter $\xi$ is a neural message function and $f_\theta$ with parameter $\theta$ is an output function. 
It takes the textual features of all tokens in a note and embeddings of code vectors as inputs and produces an interaction score between the note and each code.
To improve the computational efficiency, we take the dot product as the message function $g_\xi$.
The explicit interaction score between token $w_i$ and code $c_j$ is calculated as
\begin{equation}
	\mathbf{I}_{ij} = \mathbf{V}_{i,:}\mathbf{U}_{j,:}^{\operatorname{T}},
\end{equation}
where $\mathbf{V}_{i,:}$ is the row vector of textual features representing the $i$-th word, $\mathbf{U}_{j,:}$ is the row vector of ICD code matrix representing the $j$-th code in ICD code set.
We set $d_u = d_v$ and get the interaction matrix $\mathbf{I} \in \mathbb{R}^{m\times n}$ with dot product. 
We use a fully connected network $f_\theta$ to calculate the scores of the note-code interactions as output.
Similar to the matrix factorization formulation of language models~\cite{yang2017breaking, li2020sentence}, this dot-product interaction between notes and codes approximates the point-wise mutual information of note-code co-occurrence.

\subsection{Parameter-efficient Weight Sharing} 
The embedding injection and convolutional feature concatenation make the hidden feature high-dimensional. 
Moreover, as a result of stacking deep layers, the overall model will become cumbersome. 
Thus, we utilize a weight sharing mechanism~\cite{bai2019trellis} to decrease the number of parameters.
Specifically, we share the weights of gated CNN layers across time steps and depth through neural layers. 
This mechanism has two benefits. 
First, it can decrease the number of trainable parameters because weights across the network are tied.
Second, it provides a form of regularization to stabilize the training process.

\subsection{Objective and Training}
We formulate the ICD code assignment as a multi-label multi-class classification problem. 
We adopt the binary cross entropy loss denoted as:
\begin{equation}
\label{eq:bce}
\mathcal{L}=\sum_{i=1}^{m}\left[-y_{i} \log \left(\hat{y}_{i}\right)-\left(1-y_{i}\right) \log \left(1-\hat{y}_{i}\right)\right],
\end{equation}
where $y_{i} \in\{0,1\}$ is the ground-truth label, $\hat{y}_{i}$ is the sigmoid score for prediction, and $m$ is the number of ICD codes.
We use Adam optimizer~\cite{kingma2014adam} to train the model with backpropagation. 

\section{Experiments}
\label{sec:experiments}
In the experimental analysis on real-world datasets, we compare our proposed model with several recent strong baselines. 
Our code is available\footnote{\url{https://agit.ai/jsx/GatedCNN-NCI}}. 

\subsection{Datasets}
This paper focuses on textual discharge summaries from a hospital stay. 
Specifically, we use raw notes, ICD diagnoses, and procedures for patients from two public clinical datasets, i.e., MIMIC-II and MIMIC-III\footnote{\url{https://mimic.physionet.org}}, for experiments.  
Discharge summaries labeled with a set of ICD-9 diagnosis and procedure codes include descriptions of procedures performed by the physician, diagnosis notes, patient's medical history, and discharge instructions.

\vspace{4pt}
\noindent \textbf{MIMIC-II.}
The first dataset of clinical notes is from the Multiparameter Intelligent Monitoring in Intensive Care II (MIMIC-II) database~\cite{saeed2011multiparameter}. We follow the standard train-test split performed by Perotte et al.~\cite{perotte2014diagnosis}, where 90\% and 10\% of 22,815 non-empty discharge summaries are used for training and testing, respectively. 

\vspace{4pt}
\noindent \textbf{MIMIC-III.}
The second dataset is an updated database from Medical Information Mart for Intensive Care III (MIMIC-III) repository~\cite{johnson2016mimic}, containing patient admitted to Intensive Care Unit (ICU) at a US medical center during 2001 to 2012. 
We use the ``noteevents" table in the latest version 1.4, with 58,576 hospital admissions. 
Free-text discharge summaries in the MIMIC-III database are extracted to form the dataset with clinical text. 
The experimental evaluation considers two settings. The first one uses the full set of ICD codes.
Following Shi et al.~\cite{shi2017towards} and Mullenbach et al.~\cite{mullenbach2018explainable}, an additional experiment on the subset of MIMIC-III with the top 50 frequent labels is conducted. 
This MIMIC-III top-50 subset has a train/dev/test split with 8,066, 1,573, and 1,729 samples.

\subsection{Settings}
\label{sec:settings}

\noindent \textbf{Preprocessing}
We preprocess the textual documents following the preprocessing procedures developed by~\citet{mullenbach2018explainable} and \citet{li2020multirescnn}.
The NLTK package\furl{www.nltk.org} is utilized for tokenization, and all tokens are converted into lowercase.
All words appearing in less than three training documents were replaced with ``unk".
We truncate all documents at the length of 2500 tokens.
The word embeddings are initialized with embedding vectors pre-trained on all discharge notes with the continuous-bag-of-words (CBOW) method of $\operatorname{word2vec}$~\cite{mikolov2013distributed}.

\vspace{4pt}
\noindent \textbf{Hyper-parameters} Some standard settings follow the prior works. For example, the word embedding dimension is 100, and the dropout rate is 0.2. 
Adam optimizer~\citep{kingma2014adam} is used to optimize our model parameters. 
For the rest hyper-parameters, the random search is utilized to search the optimal settings.
The searching range or choices of specific hyper-parameters are listed in Table~\ref{tab:hyper}.
The searching interval of learning rate is $\mathopen[1\mathrm{e}^{-6}, 1\mathrm{e}^{-2}\mathopen]$. 
Besides, we optimize for kernel size, levels of residual connections, and hidden representation dimension. 

\begin{table}[h!]
\scriptsize
\caption{Range and choices of hyper-parameter search}
\begin{center}
\begin{tabular}{l | c}
\toprule
Hyper-parameters & Range/choices \\
\midrule
Learning rate & $\mathopen[1\mathrm{e}^{-6}, 1\mathrm{e}^{-2}\mathopen]$ \\
Kernel size & 2, 3, 5, 9\\
CNN levels & 1, 2, 3, 4, 5 \\
Hidden dimension & 100, 200, 300, 400, 500, 600 \\
\bottomrule
\end{tabular}
\end{center}
\label{tab:hyper}
\end{table}%

\vspace{4pt}
\noindent \textbf{Evaluation Metrics}
We use area under the receiver operating characteristic curve (AUC-ROC), F1-score, and precision at $k$ (P@$k$) for evaluation. 
We set $k=5$ for MIMIC-III subset with top-$50$ frequent codes and $k=8$ for full sets of MIMIC-II and MIMIC-III.
In the multi-label classification setting, we use two averaging strategies, i.e., micro and macro.
The macro scores are obtained by averaging the respective label-wise scores across all labels. 
Micro scores give more weight to frequent labels by considering all labels jointly. 
We run the experiments for 5 times and report the mean $\pm$ standard deviation.

\begin{table*}[h!]
\scriptsize
\centering
\setlength{\tabcolsep}{2pt}
\caption{Results on MIMIC-III with top-50 and full codes. ``-'' indicates no results reported in the original paper. \textbf{Bold} text denotes the best and \textit{italic} text denotes the second best.} \begin{center}
\begin{tabular}{lrr|rr|r|rr|rr|r}
\toprule
 \multirow{3}{4em}{Model} 	& \multicolumn{5}{c}{MIMIC-III Top-50 Codes} & \multicolumn{5}{c}{MIMIC-III Full Codes} \\
	\cline{2-11} 
	& \multicolumn{2}{c}{AUC-ROC} & \multicolumn{2}{c}{ F1 } & \multirow{2}{3em}{P@5} & \multicolumn{2}{c}{AUC-ROC} & \multicolumn{2}{c}{ F1 } & \multirow{2}{3em}{P@8}\\  
 	&Macro &Micro& Macro&Micro &  &Macro &Micro& Macro&Micro &\\
\midrule	
Bi-GRU~\cite{mullenbach2018explainable}	&	82.8	&	86.8	&	48.4	&	54.9	&	59.1	&	82.2	&	97.1	&	3.8	&	41.7	&	58.5	\\
C-MemNN~\cite{prakash2017condensed}	&	83.3	&	-	&	-	&	-	&	42.0	&	-	&	-	&	-	&	-	&	-	\\
CNN~\cite{kim2014convolutional}	&	87.6	&	90.7	&	57.6	&	62.5	&	62.0	&	80.6	&	96.9	&	4.2	&	41.9	&	58.1	\\
Attentive LSTM~\cite{shi2017towards}	&	-	&	90.0	&	-	&	53.2	&	-	&	-	&	-	&	-	&	-	&	-	\\
DR-CAML~\cite{mullenbach2018explainable}	&	88.4	&	91.6	&	57.6	&	63.3	&	61.8	&	89.7	&	98.5	&	8.6	&	52.9	&	69.0	\\
LEAM~\cite{wang2018joint}	&	88.1	&	91.2	&	54.0	&	61.9	&	61.2	&	-	&	-	&	-	&	-	&	-	\\
MultiResCNN~\cite{li2020multirescnn}		&	89.9$\pm$0.4	&	92.8$\pm$0.2	&	60.6$\pm$1.1	&	67.0$\pm$0.3	&	64.1$\pm$0.1 	& 91.0$\pm$0.2&	98.6$\pm$0.1	&	8.5$\pm$0.7	&	\textit{55.2}$\pm$0.5	&	\textit{73.4}$\pm$0.2	\\
HyperCore~\cite{cao2020hypercore}	&	89.5$\pm$0.3 	&	92.9$\pm$0.2	&	60.9$\pm$0.1	&	66.3$\pm$0.1	&	63.2$\pm$0.2 	&	\textbf{93.0}$\pm$0.1	&	\textbf{98.9}$\pm$0.5	&	\textit{9.0}$\pm$0.3	&	55.1$\pm$0.1	&	72.2$\pm$0.2	\\
\hline
GatedCNN-NCI (ours)	& \textbf{91.5}$\pm$0.3 	&	\textbf{93.8}$\pm$0.1	&	\textbf{62.9}$\pm$0.5	&	\textbf{68.6}$\pm$0.1	&	\textbf{65.3}$\pm$0.1		&	\textit{92.2}$\pm$0.2	&	\textbf{98.9}$\pm$0.3&	\textbf{9.2}$\pm$0.2	&	\textbf{56.3}$\pm$0.1	&	\textbf{73.6}$\pm$0.3\\
\bottomrule
\end{tabular}
\end{center}
\label{tab:mimic-iii}
\end{table*}%

\subsection{Baselines}
\label{sec:baselines}
We consider the following baseline models.
MultiResCNN~\citep{li2020multirescnn} and HyperCore~\cite{cao2020hypercore} are two recent strong models with the state-of-the-art performance. 
\textbf{Bi-GRU~\citep{mullenbach2018explainable}} uses a simplified gated recurrent unit with bi-direction, where last hidden representations are used for classification. 
\textbf{C-MemNN~\citep{prakash2017condensed}} introduces an iterative condensation of memory representations and utilizes external knowledge source from Wikipedia to enhance memory networks by preserving the hierarchical structure in the memory. 
\textbf{AttentiveLSTM~\citep{shi2017towards}} encodes clinical descriptions and ICD long titles jointly with character- and word-level LSTM networks and uses attention mechanism for matching important diagnosis snippets. 
\textbf{CAML~\citep{mullenbach2018explainable}} integrates CNNs and a label-wise attention mechanism to learn rich representations. It has a variant called DR-CAML that uses ICD code descriptions to regularized the loss function. 
\textbf{LEAM~\citep{wang2018joint}} encodes two channels of inputs and leverages the compatibility between word and label embeddings to calculate attention scores.
\textbf{MultiResCNN~\citep{li2020multirescnn}} combines residual learning \citep{he2016deep} and multiple channels concatenation with different convolutional filters, achieving good performance in most settings.
\textbf{HyperCore~\cite{cao2020hypercore}} utilizes hyperbolic embedding and co-graph representation with code hierarchy. It gains slightly better performance than the MultiResCNN.

\subsection{Results}
Our model performs consistently the best for frequent labels. 
First, it beats all models in the MIMIC-III subset with top-50 codes (columns 2-6 in Table~\ref{tab:mimic-iii}).
For the micro scores that give more weight to frequent labels, our model also has the best predictive metrics (columns 8\&10 in Table~\ref{tab:mimic-iii} and columns 3\&5 in Table~\ref{tab:mimic-ii-full}).
Moreover, our model is competitive also with the rest of the metrics: it consistently has the best P@k scores and at worst, the second best macro scores in all datasets.

\vspace{4pt}
\noindent \textbf{MIMIC-III (Top-50 Codes)}
The first experiment uses the MIMIC-III subset with top-50 codes, showing models' performance on predicting the frequent diagnosis.  
The results in Table~\ref{tab:mimic-iii} show that our model outperforms all the baselines in all the evaluation metrics. 
Significantly, our model gains a higher macro F1-score by 2\% and micro F1-score by 1.6\% than the state of the art.

\vspace{4pt}
\noindent \textbf{MIMIC-III (Full Codes)}
We then run our model on the MIMIC-III dataset with full codes. 
Our model outperforms most baselines, gaining the best scores in macro AUC-ROC, macro F1, micro F1, and precision@8.
For the macro AUC-ROC, our model is ranked at the second place.

\vspace{4pt}
\noindent \textbf{MIMIC-II (Full Codes)}
In the third dataset of MIMIC-II, we also predict the full set of ICD-9 codes. 
Our model achieves predictive performance on par with two recent strong baselines of MultiResCNN and HyperCore. 
We gain the best scores in micro AUC-ROC, micro F1-score, and P@8. 
Macro AUC-ROC and F1 scores of our model are the second best of the models compared. 

\begin{table}[h!]
\scriptsize
\setlength{\tabcolsep}{2pt}
\caption{Results on MIMIC-II full codes. \textbf{Bold} text denotes the best and \textit{italic} text denotes the second best.}
\begin{center}
\begin{tabular}{lrr|rr|r}
\toprule
 \multirow{2}{4em}{Model} & \multicolumn{2}{c}{AUC-ROC} & \multicolumn{2}{c}{ F1 } & \multirow{2}{3em}{P@8} \\  
 	&Macro &Micro& Macro&Micro &  \\
\midrule			
CNN 	&	74.2	&	94.1	&	3.0	&	33.2	&	38.8	\\
Bi-GRU 	&	78.0	&	95.4	&	2.4	&	35.9	&	42.0	\\
DR-CAML 	&	82.6	&	96.6	&	4.9	&	45.7	&	51.5	\\
MultiResCNN 	&	85.0$\pm$0.2	&	96.8$\pm$0.1&	5.2$\pm$0.2	&	46.4$\pm$0.2	&	\textit{54.4}$\pm$0.7\\
HyperCore 	&	\textbf{88.5}$\pm$0.1 &	\textit{97.1}$\pm$0.4	&	\textbf{7.0}$\pm$0.2	&	\textit{47.0}$\pm$0.3	&	53.7$\pm$0.3	\\
\hline
GatedCNN-NCI	&	\textit{87.2}$\pm$0.3	 &	\textbf{97.2}$\pm$0.1	&	\textit{6.4}$\pm$0.3	&	\textbf{47.3}$\pm$0.2	&	\textbf{54.5}$\pm$0.4	\\
\bottomrule
\end{tabular}
\end{center}
\label{tab:mimic-ii-full}
\end{table}%

\subsection{Comparison with BERT}
Bidirectional Encoder Representations from Transformers (BERT)~\cite{devlin2019bert} has revolutionized the NLP community recently. 
The pre-trained language model has been applied to different downstream NLP tasks. 
We compare our model's performance with the BERT model and a domain-specific variant, i.e., ClinicalBERT~\cite{alsentzer2019publicly} pre-trained on the clinical text of MIMIC-III.
For the BERT model, we use the uncased BERT-base with a hidden dimension of 768. 
Because these two BERT models require the configuration of the maximum sequence length of 512, we truncate the text sequence for our model to ensure a fair comparison. 
BERT models have two special tokens, i.e., \texttt{[CLS]} and \texttt{[SEP]}. 
Thus, we truncate clinical notes with a length of 510. 
We use Huggingface's transformer framework\footnote{\url{https://github.com/huggingface/transformers}} when implementing these two models. 
The results in Table~\ref{tab:bert} show that pretraining the language model with domain data improves the performance, and our model has better performance in most evaluation metrics. 

\begin{table}[h]
\scriptsize
\caption{Comparison with BERT and ClinicalBERT using the MIMIC-III top-50 code dataset with sequence length truncated at 510.}
\begin{center}
\begin{tabular}{lrr|rr|r}
\toprule
 \multirow{2}{4em}{Model} & \multicolumn{2}{c}{AUC-ROC} & \multicolumn{2}{c}{ F1 } & \multirow{2}{2em}{P@5}\\  
 	&Macro &Micro& Macro&Micro &  \\
\midrule
BERT-base	&	80.6	&	85.2	&	43.3	&	53.2	&	53.3	\\
ClinicalBERT	&	81.0	&	85.6	&	\textbf{43.9}	&	54.3	&	54.5	\\
GatedCNN-NCI	&	\textbf{83.7}	&	\textbf{87.7	}&	42.9	&	\textbf{54.4} 	&	\textbf{56.6}	\\
\bottomrule
\end{tabular}
\end{center}
\label{tab:bert}
\end{table}%

\subsection{Model Size}
\label{sec:efficiency}
We compare the number of trainable parameters (Table~\ref{tab:efficiency}) of our model with two models with qualified performance, i.e., CAML~\citep{mullenbach2018explainable} and MultiResCNN~\citep{li2020multirescnn}.
HyperCore~\cite{cao2020hypercore} didn't publish the code or provide the values of all hyperparameters. Thus, we omit it in this comparison. 
Our proposed model is more efficient than the MultiResCNN in terms of the number of trainable parameters. 
The CAML model has the fewest parameters but performs poorly in prediction. 
Our model has a much better predictive performance than the CAML model, with only a slight increase in model size.

\begin{table}[!h]
\scriptsize
\caption{Number of trainable parameters}
\begin{center}
\begin{tabular}{l r } %
\toprule
Model	&	num. params. 	\\
\midrule
CAML~\citep{mullenbach2018explainable}	&	6.2M	\\
MultiResCNN~\citep{li2020multirescnn}	&	11.9M	\\
ClinicalBERT~\citep{alsentzer2019publicly} & 113.8M \\
GatedCNN-NCI (Ours)	&	7.6M	\\
\bottomrule
\end{tabular}
\end{center}
\label{tab:efficiency}
\end{table}%

\subsection{Ablation Study}
We further conduct an ablation study the investigate the effectiveness of different components of our proposed model. 
We evaluate two variants by removing two critical components of the proposed model.
The first variant without NCI replaces the note-code interaction with max-pooling and linear projection.
The second variant removes the gating mechanism that controls the information prorogation over the CNN layers.
Table~\ref{tab:ablation} compares the experimental results on the MIMIC-III subset with top-50 codes.
The performance drops to some extent after removing these two modules, which shows the effectiveness of our proposed architectures. 
Moreover, the note-code interaction module has slightly more contribution than the gating mechanism.
Possible explanations are that the explicit interaction perseveres the semantics of medical codes well and captures the relation between codes and notes in the embedding space. 

\begin{table}[h]
\scriptsize
\caption{Ablation study}
\begin{center}
\begin{tabular}{lrr|rr|r}
\toprule
 \multirow{2}{4em}{Model} & \multicolumn{2}{c}{AUC-ROC} & \multicolumn{2}{c}{ F1 } & \\  
 	&Macro &Micro& Macro&Micro & P@5 \\
\midrule
GatedCNN-NCI	&	91.5	&	93.8	&	62.9	&	68.6	&	65.3	\\
without NCI	&	90.1	&	92.7	&	61.4	&	67.2	&	63.9	\\
without gating	&	90.0	&	92.0	&	60.2	&	66.9	&	63.7	\\	
\bottomrule
\end{tabular}
\end{center}
\label{tab:ablation}
\end{table}%

\subsection{Case Study}

We conduct a case study to interpret an example prediction. Table \ref{tab:case} shows the predictions for a clinical note of a patient with cardiovascular diseases and diabetes. The patient also had `dyspnea on exertion' as a symptom caused by either pneumonia or cardiac diseases. Our model and MultiResCNN predict the correct diagnosis codes: coronary atherosclerosis (ICD code: 414.01), hypertension (401.9), and diabetes (250.00). When predicting procedure codes, MultiResCNN is confused by dyspnea on exertion and incorrectly predicts pneumonia-related treatments: endotracheal intubation (96.04) and invasive mechanical ventilation (96.71). Our model correctly predicts a cardiac catheterization procedure and diagnostic interventions of heart surgery (39.61) and coronary artery bypass (36.15).     
Hence, our model is not misled by the ambiguous interpretation for dyspnea on exertion but learns the correct cardiac-related context, consistent with the rest of the note.

\begin{table}[h!]
\scriptsize
\caption{Case study on a clinical note with cardiac-related diseases (\textbf{bold}, in \hlgreen{green}). Dyspnea on exertion (\textit{italic}, in \hlred{red}) can be caused by cardiac- or pneumonia-related diseases.}
\centering
\setlength{\tabcolsep}{2pt}
\renewcommand{\arraystretch}{1}
\linespread{0.1}
\begin{tabular}{p{60pt}|p{60pt}|p{70pt}}
\toprule
Clinical note &	 \multicolumn{2}{p{160pt}}{old male with multiple \textbf{\hlgreen{cardiac risk factors}} and \textit{\hlred{dyspnea on exertion}} $\dots$, he then underwent further workup which included \textbf{\hlgreen{a cardiac catheterization}} that revealed significant \textbf{\hlgreen{coronary artery disease}}. he was then transferred for surgical evaluation".} 
\\
\midrule
Prediction & Procedure codes & Diagnosis codes \\
\midrule
Gold ICD codes & \textbf{\hlgreen{36.15}}; ~~\textbf{\hlgreen{39.61}};& ~~401.9; ~~414.01; ~~250.00	\\
MultiResCNN &  \textit{\hlred{96.04}}; ~~\textit{\hlred{96.71}};& ~~401.9; ~~414.01; ~~250.00 	\\
GatedCNN-NCI & \textbf{\hlgreen{36.15}}; ~~\textbf{\hlgreen{39.61}};& ~~401.9; ~~414.01; ~~250.00 \\
\bottomrule	
\end{tabular}

\label{tab:case}
\end{table}

\section{Conclusion}
\label{sec:conclusion}
Medical code assignment from clinical notes is a fundamental task for healthcare information systems and diagnosis decision support. 
This paper proposes a novel framework with gated convolutional neural networks and note-code message passing mechanism for automated medical code assignment.
Our solution can learn meaningful features from lengthy clinical documents and effectively control the deep propagation of information flow. 
Moreover, the message passing mechanism can enhance the ICD code space's semantics and model the note-code interaction to improve medical code prediction.
Experiments show the effectiveness of our proposed method.
 
\section*{Acknowledgments}
This work was supported by the Academy of Finland (grant 336033) and EU H2020 (grant 101016775).
We acknowledge the computational resources provided by the Aalto Science-IT project.
The authors wish to acknowledge CSC - IT Center for Science, Finland, for computational resources.

\bibliographystyle{acl_natbib}
\bibliography{mca}

\end{document}